\documentclass[letterpaper, 10 pt, conference]{ieeeconf}  
\IEEEoverridecommandlockouts                              
\usepackage{graphicx}
\usepackage{booktabs}
\usepackage{amssymb}
\usepackage{makecell}
\usepackage{tabularx}
\usepackage{amsmath}
\usepackage{titlesec}
\let\labelindent\relax
\usepackage{enumitem}
\usepackage{caption}
\usepackage{float}
\usepackage[font=small]{caption}
\usepackage{titlesec}
\usepackage{subcaption}
\usepackage[table]{xcolor}
\usepackage{pifont}
\usepackage{wrapfig}
\usepackage{tabularx}
\usepackage[most]{tcolorbox}
\usepackage{xcolor}
\usepackage{algorithm}
\usepackage{algpseudocode}
\usepackage{placeins}
\usepackage{balance}

\newtcolorbox{promptbox}[1]{
    colback=gray!4,
    colframe=gray!55,
    title=#1,
    fonttitle=\bfseries,
    coltitle=black,
    boxrule=0.4pt,
    arc=1.5mm,
    left=1mm,
    right=1mm,
    top=1mm,
    bottom=1mm,
    before skip=0.6em,
    after skip=0.6em,
    breakable
}

\newcolumntype{C}{>{\centering\arraybackslash}X}

\newcommand{\cmark}{\textcolor{green!60!black}{\ding{51}}}
\newcommand{\xmark}{\textcolor{red!80!black}{\ding{55}}}

\titlespacing*{\section}
{0pt}{0.5ex plus 0.1ex minus 0.1ex}{0.3ex plus 0.1ex}

\titlespacing*{\subsection}
{0pt}{0.4ex plus 0.1ex minus 0.1ex}{0.2ex plus 0.1ex}

\titlespacing*{\subsubsection}
{0pt}{0.3ex plus 0.1ex minus 0.1ex}{0.15ex plus 0.05ex}

\setlist[itemize]{topsep=2pt, itemsep=1pt, parsep=0pt, leftmargin=*}
\setlist[enumerate]{topsep=2pt, itemsep=1pt, parsep=0pt, leftmargin=*}

\AtBeginDocument{%
  \setlength{\abovedisplayskip}{3pt}
  \setlength{\belowdisplayskip}{3pt}
  \setlength{\abovedisplayshortskip}{1pt}
  \setlength{\belowdisplayshortskip}{1pt}
}

\title{\LARGE \bf
AgenticRL: Agentic Reinforcement Learning with Self-Refinement for Complex UAV Navigation%
}

\author{Roohan Ahmed Khan, Yasheerah Yaqoot, Amir Atef Habel,
Muhammad Ahsan Mustafa, Dzmitry Tsetserukou
}
\begin{document}

\maketitle
\thispagestyle{empty}
\pagestyle{empty}

\begin{abstract}

Deep reinforcement learning enables autonomous robots to learn complex navigation tasks, but still relies heavily on time consuming manual reward design and fine tuning. Existing automated reward generation and refinement methods reduce this effort, yet often lack task-level behavioral diagnosis for directing subsequent reward revisions. We introduce AgenticRL, a multimodal closed loop framework in which role-specialized agents generate executable rewards, diagnose failures of the resulting policies, formulate targeted refinement instructions, and regenerate improved rewards. Before training, a task grounding stage automatically selects a compatible action profile, together with its observation and reward interfaces. Each generated reward is used to train a policy using Proximal Policy Optimization (PPO), which is subsequently evaluated under randomized conditions. Task-level behavioral, geometric, and safety measurements are organized into a structured diagnosis packet and jointly analyzed with the current reward code, task specification, behavioral summary, and visual scene context. Unlike one-shot reward generation, human-guided refinement, or broad candidate search, AgenticRL uses automated diagnosis of the behavior induced by a reward to direct its next revision. We evaluate the framework across eight UAV tasks covering navigation, obstacle interaction, trajectory tracking, agile manoeuvres, and cluttered flight. Under the reported comparative evaluation, AgenticRL achieves success rates of $100\%$ in racing and $88\%$ in cluttered navigation, exceeding the strongest Eureka and Text2Reward baselines, respectively. Reward refinement increases mean simulation success from $37.2\%$ to $96.4\%$, while the resulting policies achieve a collective real-world success rate of $90.0\%$ and a sim-to-real accuracy of $93.4\%$.
\end{abstract}

\section{Introduction}

Unmanned aerial vehicles (UAVs) are increasingly used in inspection, search and rescue, agriculture, and navigation in cluttered or GPS denied environments. Reinforcement learning (RL) has shown strong potential for UAV navigation and control~\cite{joshi2023simtorealdeepreinforcementlearning, wang2024drluav, multi-stage-drl}, but its performance remains highly dependent on reward design. For complex robotic tasks, manually translating desired behaviors into numerical objectives is difficult, and poorly designed rewards can lead to unstable learning, unsafe behavior, incomplete task execution, or reward exploitation. Existing approaches based on demonstrations, preferences, or human feedback~\cite{christiano2017deep, sadigh2017active, active-preference-gaussian} can reduce manual reward engineering, but often require additional supervision or task-specific data.

\begin{figure}[t!]
\centering
\includegraphics[width=\linewidth]{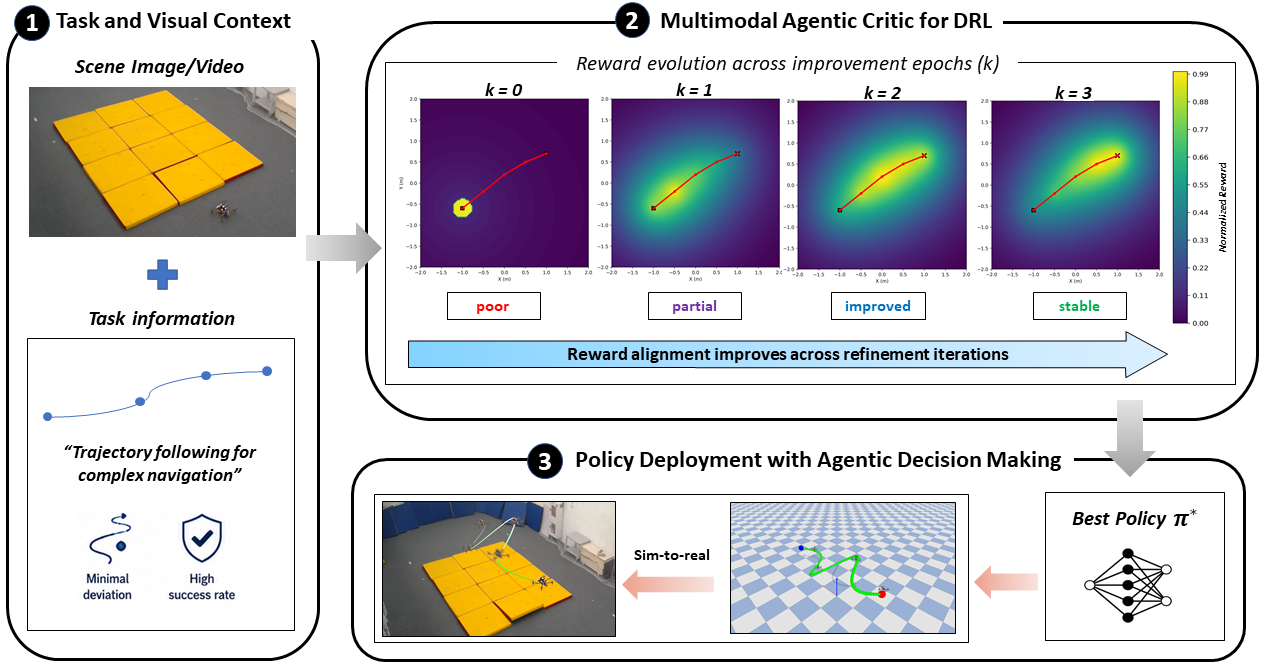}
\captionof{figure}{AgenticRL framework for reward refinement and policy deployment. A VLM agent generates task-specific rewards, evaluates trained policies, and acts as an agentic critic by diagnosing policy behavior and refining rewards across iterative improvement epochs. The final refined reward produces the best policy, which is selected and deployed through agentic decision making for sim-to-real drone navigation.}
\label{fig:title_figure}
\end{figure}

Recent advances in large language models (LLMs) and vision language models (VLMs) have enabled automated policy and reward generation. Prior work has explored language conditioned policy synthesis~\cite{liang2022code}, reward generation from task descriptions~\cite{xie2024text2reward}, and iterative LLM-based reward optimization~\cite{ma2024eureka}. However, generating a plausible reward function does not guarantee that the resulting RL policy will exhibit the intended behavior. Existing approaches commonly rely on one-shot generation, reward statistics, or candidate search, rather than explicitly diagnosing task-level policy failures and using behavioral evidence to guide subsequent reward revisions.

To address this limitation, we propose \textbf{AgenticRL} shown in Fig.~\ref{fig:title_figure}, a multimodal closed loop framework for behavior-grounded reward refinement in UAV reinforcement learning. Given a natural language task specification, scene image, and observation definition, AgenticRL generates an executable reward function and trains a policy in simulation. The trained policy is then evaluated under randomized task conditions to construct a structured diagnosis packet containing task-specific behavioral and geometric evidence, such as failure stage, collisions, dynamic goal error, manoeuvre completion, obstacle clearance, goal progress, and recovery behavior. A multimodal analysis agent jointly reasons over this evidence, the current reward code, task specification, behavioral summary, and scene context to identify likely reward design deficiencies and generate a targeted refinement for the next iteration. Thus, AgenticRL closes the loop between the reward function and the behavior it actually induces, rather than relying primarily on one-shot generation or broad candidate sampling. The main contributions of our work are:
\begin{itemize}
    \item A multimodal AgenticRL framework with role-specialized agents for executable reward generation, behavioral diagnosis, and targeted reward refinement in a closed loop.

    \item An automated task grounding and configuration mechanism that maps each task to compatible simulation, observation, reward, controller, and action-profile interfaces before training.

    \item A scenario registry that supports multimodal scenario recognition and automatic selection of trained policies for deployment.

    \item Evaluation across eight UAV tasks with distinct behaviors and action profiles, including validation on a real world quadrotor platform.
\end{itemize}


\section{Related Works}

\subsection{UAV Reinforcement Learning and Reward Design}
Reinforcement learning has been widely studied for UAV navigation, obstacle avoidance, trajectory tracking, and sim-to-real control~\cite{joshi2023simtorealdeepreinforcementlearning, wang2024drluav, multi-stage-drl}. Prior work has shown that reward shaping can improve learning efficiency and policy performance~\cite{millan2022reinforcement, ng1999policy, devidze2021explicable}. However, these approaches still rely heavily on manually designed rewards, hand-tuned coefficients, and task-specific engineering, which becomes increasingly difficult for UAV tasks involving safety, stability, and collision avoidance.

\subsection{Language and Vision Language Models for Learning}
Recent works have explored the use of LLMs and VLMs for robot learning and reward generation. LLMs have been used for robot policy synthesis~\cite{liang2022code}, proxy reward modeling~\cite{kwon2023reward}, executable reward generation from language~\cite{xie2024text2reward}, and learning rewards from natural language feedback or demonstrations~\cite{words-to-rewards, zhang2025rewind}. Similarly, VLM-based methods use visual grounding to generate or evaluate robotic rewards~\cite{code-as-reward, lee2026roborewardgeneralpurposevisionlanguagereward}. While these approaches reduce manual reward engineering, many rely on demonstrations, symbolic abstractions, human feedback, or one-shot reward generation rather than task-level diagnosis of the behavior induced by a trained policy.

\subsection{Agentic Learning Pipelines}
Recent work on agentic reinforcement learning shows that LLM-based agents can improve through feedback, reasoning, tool use, and iterative refinement~\cite{zhang2026landscape, liu2026implicit, zhang2025agentrl, wang2025agentrlvr, fan2026reasoning}. Other studies investigate scalable agentic RL systems and tool-augmented reward models for long-horizon tasks~\cite{hu2025openrewardlearningrewardlongform, li2026literesearcherscalableagenticrl}. These methods primarily target text-based or reasoning-oriented environments rather than embodied UAV learning.

Among related automated RL frameworks, Agents Trainer~\cite{agents-trainer} automates parts of the MARL pipeline for drone swarms using specialized agents for reward generation and evaluation. However, it targets multi-agent training automation rather than multimodal reward refinement driven by task-level policy behavior. AgenticRL instead combines multimodal task grounding, executable reward generation, structured behavioral diagnosis, and iterative reward refinement for UAV policy learning, with validation across multiple simulation and real-world scenarios.

Table~\ref{tab:related_comparison} summarizes the main differences between AgenticRL and the most relevant existing frameworks. Prior work has made substantial progress in robot policy code generation, language guided reward synthesis, VLM-based reward generation, and agentic RL pipelines. However, existing methods usually address only part of the full problem. Some generate policy code directly, some generate rewards from text, some use visual inputs only for reward verification, and others automate MARL pipelines without explicit visual scene conditioning. AgenticRL combines these directions by generating reward code from language and scene context, training UAV policies in simulation, diagnosing policy behavior, refining rewards in a closed loop, and validating the learned policies on a real UAV.
\begin{table}[t]
\vspace{-8pt}
\centering
\caption{Comparison with Related Frameworks.}
\label{tab:related_comparison}
\scriptsize
\setlength{\tabcolsep}{3.2pt}
\renewcommand{\arraystretch}{1.10}

\resizebox{\columnwidth}{!}{%
\begin{tabular}{lccccc}
\toprule
\textbf{Method} &
\makecell{\textbf{Reward}\\\textbf{Code}} &
\makecell{\textbf{Visual}\\\textbf{Input}} &
\makecell{\textbf{Closed loop}\\\textbf{Refinement}} &
\makecell{\textbf{Full Task}\\\textbf{Diagnosis}} &
\makecell{\textbf{UAV}\\\textbf{Implementation}} \\
\midrule
\rowcolor{gray!14}
Code as Policies & \xmark & \xmark & \xmark & \xmark & \xmark \\
Text2Reward & \cmark & \xmark & Human-guided & Limited Logs & \xmark \\
\rowcolor{gray!14}
Eureka & \cmark & \xmark & \cmark & Reward Statistics & \xmark \\
Code as Reward & \cmark & \cmark & \xmark & Subtasks & \xmark \\
Agents Trainer & \cmark & \xmark & \cmark & Error Logs only & \cmark \\
\midrule
\rowcolor{gray!20}
\textbf{AgenticRL} & \cmark & \cmark & \cmark & \cmark & \cmark \\
\bottomrule
\end{tabular}%
}

\vspace{-8pt}
\end{table}

\begin{figure*}[h!]
    \centering
    \includegraphics[width=0.95\textwidth]{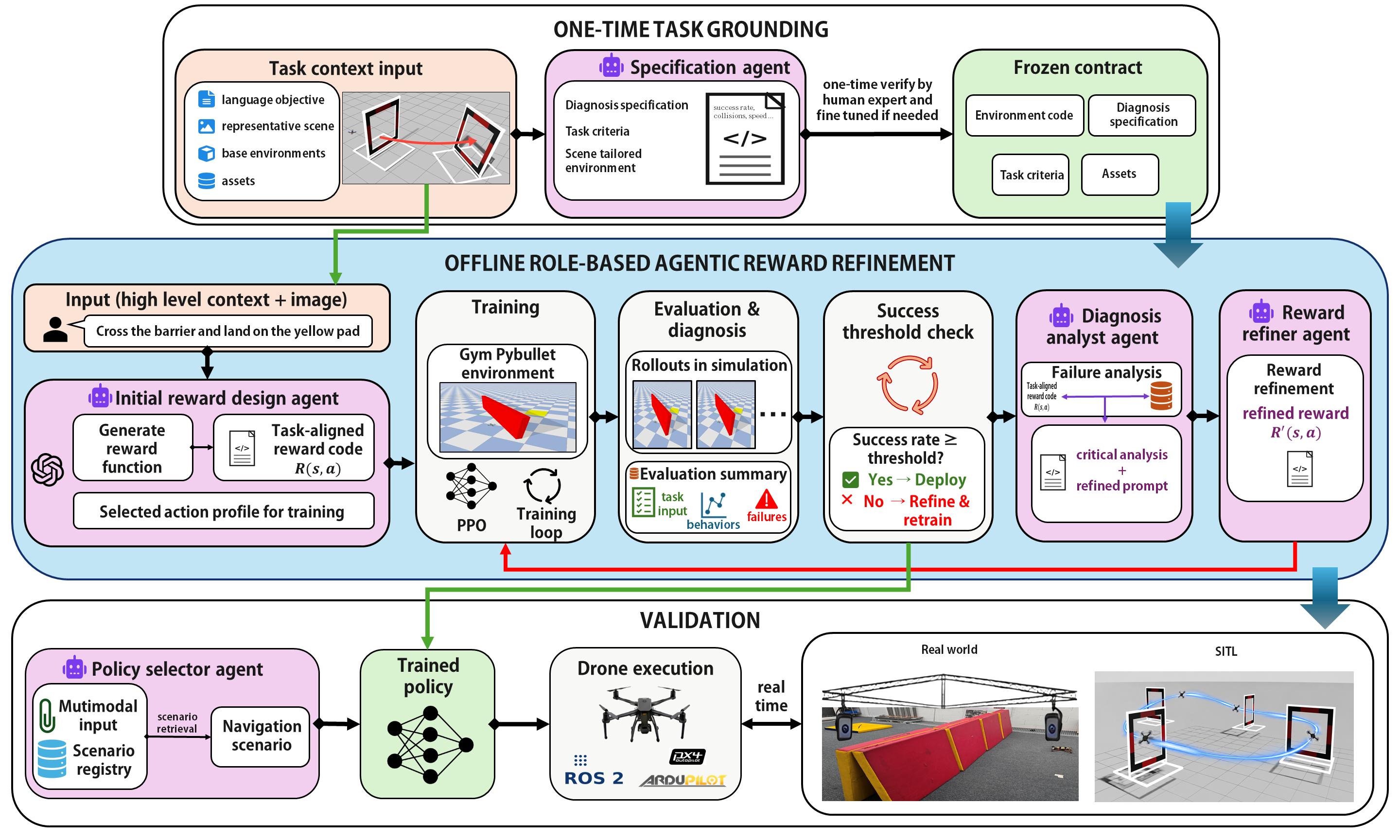}
    \caption{System architecture of AgenticRL. A one-time task grounding stage defines the task specification and required evaluation criteria, followed by an offline role-specialized loop for reward generation, policy training, behavioral diagnosis, and iterative refinement. The best retained policy is then selected for deployment through multimodal scenario identification.}
    \label{fig:system_architecture}
    \vspace{-0.2cm}
\end{figure*}

\section{System Overview}
\label{sec:sys-overview}

\subsection{Problem Formulation}
We consider the problem of training reinforcement learning policies from high level multimodal task specifications. Each task is defined by a natural language instruction $l_0$, visual scene context $I$, and an environment observation specification $\mathcal{O}$. The visual context provides task relevant scene information, while $\mathcal{O}$ defines the state variables available to the policy and reward function. Each task is formulated as a Markov decision process (MDP):
\begin{equation}
\mathcal{M} = (\mathcal{S}, \mathcal{A}, \mathcal{P}, R, \gamma),
\end{equation}
where $\mathcal{S}$ denotes the state space, $\mathcal{A}$ is the continuous action space, $\mathcal{P}$ is the transition dynamics, $R$ is the reward function, and $\gamma$ is the discount factor. The policy $\pi_{\theta}(a_t \mid s_t)$ maps observation $s_t \in \mathcal{S}$ at time step $t$ to action $a_t \in \mathcal{A}$, represented as either a velocity command or a collective-thrust and body-rate command depending on the scenario. The observation space contains the UAV state together with task-specific quantities such as gate poses, obstacle locations, landing targets, barrier information, or trajectory waypoints.

The reward function $R$ is not explicitly available and must be inferred from the multimodal task specification. Since a semantically plausible reward may still induce incomplete, unsafe, or unintended behavior, AgenticRL first generates an executable initial reward:
\begin{equation}
R_0 = \mathcal{G}_{\phi}(l_0, I, \mathcal{O}),
\label{eq:initial_reward}
\end{equation}
where $\mathcal{G}_{\phi}$ denotes the VLM agent. The generated reward is then used to optimize the policy objective:
\begin{equation}
\theta^{*} =
\arg\max_{\theta}
\mathbb{E}_{\tau \sim \pi_{\theta}}
\left[
\sum_{t=0}^{T}
\gamma^t R_k(s_t,a_t)
\right].
\end{equation}
AgenticRL subsequently evaluates the policy behavior induced by $R_k$ and uses this evidence to guide reward refinement.

\subsection{AgenticRL Framework}
\label{sec:agenticrl_framework}
The overall architecture is shown in Fig.~\ref{fig:system_architecture}. AgenticRL consists of a one-time task grounding stage followed by an offline role-based reward refinement loop. Task grounding defines the language objective, representative scene, environment assets, diagnosis specification, and task criteria, which are fixed before iterative training begins. The refinement loop then performs reward generation, policy training, randomized evaluation, behavioral diagnosis, and reward regeneration until the success criterion is satisfied.

Given the multimodal task specification $(l_0,I,\mathcal{O})$, the initial reward design agent generates an executable Python reward function $R_0$ containing task progress, safety, and completion terms derived from the available observations. The reward is integrated into a customized UAV simulation environment~\cite{gympybyllet} and used to train a Proximal Policy Optimization (PPO)~\cite{ppo} policy. PPO is adopted due to its widespread use in continuous UAV control and drone-racing applications~\cite{championleveldrone, autonomousdroneracing}.

After training, the policy is evaluated over randomized simulation episodes using task-specific behavioral and geometric metrics. These measurements are organized into a diagnosis packet $D_k$ and a derived position--behavior summary that capture where and under what conditions the policy fails. Fig.~\ref{fig:diagnosis_example} shows an abbreviated example from the flip-traversal task, including the original diagnosis packet excerpt, the derived behavioral summary, and the resulting structured analyzer output.

\begin{figure*}[t]
\centering
\begin{minipage}[c]{0.285\textwidth}
\begin{tcolorbox}[
    title={\scriptsize Diagnosis packet $D_k$ (excerpt)},
    fonttitle=\bfseries,
    colback=blue!3,
    colframe=blue!45!black,
    boxrule=0.4pt,
    arc=1mm,
    left=0.8mm,right=0.8mm,top=0.6mm,bottom=0.6mm,
    before skip=0pt,after skip=0pt]
\ttfamily\tiny
model\_name: PPO-2026.08.28\_21.25.44\\
aggregate\_metrics:\\
\hspace*{1em}num\_scenarios: 100\\
\hspace*{1em}success\_rate: 0.00\\
\hspace*{1em}lower\_gate\_pass\_rate: 1.00\\
\hspace*{1em}flip\_completion\_rate: 0.00\\
\hspace*{1em}upper\_gate\_pass\_rate: 0.00\\
\hspace*{1em}avg\_max\_flip\_rotation\_deg: 269.8\\
\hspace*{1em}avg\_final\_vertical\_speed: 5.65\\
failure\_breakdown:\\
\hspace*{1em}lower\_gate\_passed\_flip\_incomplete: 100
\end{tcolorbox}
\end{minipage}
\hfill
\begin{minipage}[c]{0.035\textwidth}
\centering $\Longrightarrow$
\end{minipage}
\hfill
\begin{minipage}[c]{0.275\textwidth}
\begin{tcolorbox}[
    title={\scriptsize Position--behavior summary},
    fonttitle=\bfseries,
    colback=orange!4,
    colframe=orange!55!black,
    boxrule=0.4pt,
    arc=1mm,
    left=0.8mm,right=0.8mm,top=0.6mm,bottom=0.6mm,
    before skip=0pt,after skip=0pt]
\ttfamily\tiny
failure\_type\_summary:\\
\hspace*{1em}crossed\_gate\_plane:\\
\hspace*{2em}count: 100\\
\hspace*{2em}success\_rate: 0.00\\
\hspace*{2em}collision\_rate: 0.00\\[1pt]
grouped\_behavior:\\
\hspace*{1em}target\_dist\_bucket:\\
\hspace*{2em}group: dist\_far\\
\hspace*{2em}count: 100\\
\hspace*{2em}success\_rate: 0.00\\
\hspace*{2em}collision\_rate: 0.00
\end{tcolorbox}
\end{minipage}
\hfill
\begin{minipage}[c]{0.035\textwidth}
\centering $\Longrightarrow$
\end{minipage}
\hfill
\begin{minipage}[c]{0.315\textwidth}
\begin{tcolorbox}[
    title={\scriptsize Structured analyzer output (excerpt)},
    fonttitle=\bfseries,
    colback=green!3,
    colframe=green!45!black,
    boxrule=0.4pt,
    arc=1mm,
    left=0.8mm,right=0.8mm,top=0.6mm,bottom=0.6mm,
    before skip=0pt,after skip=0pt]
\ttfamily\tiny
dominant\_failure\_mode:\\
\hspace*{1em}three-quarter flip; out-of-bounds climb\\
suspected\_root\_cause:\\
\hspace*{1em}late pitch braking; spin/climb terms\\
\hspace*{1em}dominate task completion\\
evidence\_summary:\\
\hspace*{1em}lower gate 100\%; flip/upper gate 0\%\\
what\_should\_be\_done:\\
\hspace*{1em}taper pitch at progress 0.55--0.70;\\
\hspace*{1em}bound speed; broaden upper-gate guidance\\
output: refined\_prompt
\end{tcolorbox}
\end{minipage}

\vspace{2pt}


\caption{Flip-traversal refinement example showing the diagnosis packet, position-conditioned behavior summary, and multimodal analyzer output used to guide reward refinement.}
\label{fig:diagnosis_example}
\end{figure*}

The diagnosis analyst agent uses this behavioral evidence, together with the current reward, task specification, and visual scene context, to identify likely reward design deficiencies and generate targeted refinement guidance:
\begin{equation}
p_{k}^{\mathrm{ref}} = \mathcal{H}_{\phi}(D_k),
\label{eq:ref_prompt}
\end{equation}
where $\mathcal{H}_{\phi}$ denotes the diagnosis and refinement prompt operator. The resulting guidance is incorporated into the task context:
\begin{equation}
l_{k+1}
=
\Psi_{\phi}
\left(
l_k,\;
p_k^{\mathrm{ref}}
\right),
\label{eq:prompt_update}
\end{equation}
where $\Psi_{\phi}$ denotes the prompt update operator. The updated context is then passed to the reward refiner agent together with the previous reward and multimodal task information:
\begin{equation}
R_{k+1}
=
\mathcal{G}_{\phi}
\left(
R_k,\;
l_{k+1},\;
I,\;
\mathcal{O}
\right).
\label{eq:refined_reward}
\end{equation}

The refined reward is retrained and re-evaluated, closing the behavior-grounded refinement loop. If the success criterion is satisfied, the best retained policy is selected for deployment; otherwise, the diagnosis and refinement cycle continues until the maximum given iterations. 

\section{Experimental Setup}
\label{sec:exp-setup}

\begin{figure*}[h!]
    \centering
    \includegraphics[width=0.92\textwidth]{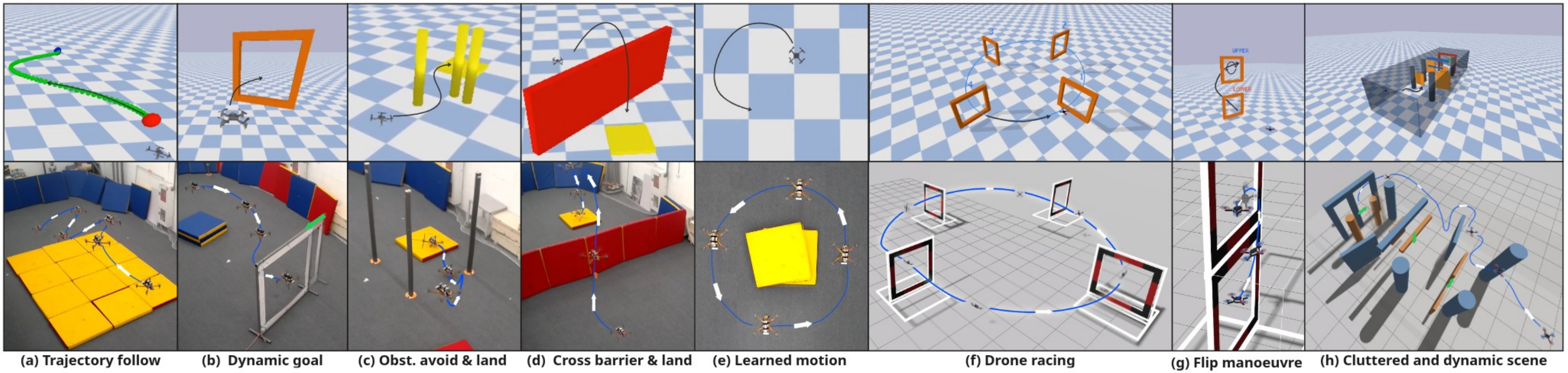}
    \caption{Overview of the eight UAV evaluation scenarios. The top row shows the corresponding PyBullet training environments, while the bottom row shows representative Gazebo or real world realizations where applicable.}
    \label{fig:uav_scenarios}
    \vspace{-0.3cm}
\end{figure*}

AgenticRL is evaluated on eight UAV tasks spanning navigation, obstacle interaction, trajectory tracking, motion generation, agile manoeuvring, and cluttered flight. Fig.~\ref{fig:uav_scenarios} provides a visual overview of all scenarios in simulation and, where applicable, Gazebo or real world deployment.

\subsection{UAV Task Scenarios}
\label{sec:uav-scenarios}

Five scenarios use linear-velocity actions: dynamic goal, obstacle avoidance and landing, trajectory following, barrier crossing and landing, and circular motion generation. In the motion-generation scenario, the policy receives only the UAV state, without a reference trajectory, and learns to produce the desired circular flight behavior directly from the task specification. Three additional agile scenarios, including racing, flip traversal, and cluttered navigation, use collective thrust and body-rate commands. These tasks evaluate the reward design workflow across different UAV behaviors and control abstractions; policy transfer between the two action profiles is not assumed.

All policies are trained using PPO with fully connected networks and $\tanh$ activations. We use 16 parallel environments, rollout length 2048, minibatch size 256, and clipping parameter $\epsilon=0.2$. Scenario-specific training budgets and hyperparameters are summarized in Table~\ref{tab:training_settings}. For comparative reward design experiments, all methods use identical policy-optimization settings and a fixed PPO training seed within each scenario. Each generated reward candidate is trained once before evaluation.
\begin{table*}[h!]
\centering
\caption{Scenario-specific PPO training settings across the two UAV action profiles.}
\label{tab:training_settings}
\scriptsize
\setlength{\tabcolsep}{6pt}
\renewcommand{\arraystretch}{1.12}
\resizebox{\textwidth}{!}{%
\begin{tabular}{lcccccc}
\toprule
\textbf{Scenario} &
\makecell{\textbf{Action profile}} &
\makecell{\textbf{Steps per reward} \\ \textbf{(million)}} &
\makecell{\textbf{Learning rate}} &
$\gamma$ &
\makecell{\textbf{Entropy coefficient}} &
\makecell{\textbf{Episode horizon (s)}} \\
\midrule
Dynamic goal          & Linear velocities     & 25 & $1\!\times\!10^{-5}$ & 0.990 & $0.10\!\rightarrow\!0.001$  & 25 \\
Obstacle avoidance and landing & Linear velocities     & 50 & $1\!\times\!10^{-5}$ & 0.990 & $0.10\!\rightarrow\!0.001$  & 25 \\
Trajectory following           & Linear velocities     & 50 & $1\!\times\!10^{-5}$ & 0.990 & $0.10\!\rightarrow\!0.001$  & 25 \\
Barrier crossing and landing   & Linear velocities     & 70 & $1\!\times\!10^{-5}$ & 0.990 & $0.10\!\rightarrow\!0.001$  & 25 \\
Circular motion generation     & Linear velocities     & 70 & $1\!\times\!10^{-5}$ & 0.990 & $0.10\!\rightarrow\!0.001$  & 25 \\
\midrule
Racing scene                   & Body rates and thrust & 10 & $1\!\times\!10^{-5}$ & 0.990 & $0.010\!\rightarrow\!0.001$ & 12 \\
Flip traversal       & Body rates and thrust & 15 & $1\!\times\!10^{-5}$ & 0.995 & $0.020\!\rightarrow\!0.002$ & 15 \\
Cluttered scene                & Body rates and thrust & 15 & $2\!\times\!10^{-5}$ & 0.995 & $0.015$                      & 20 \\
\bottomrule
\end{tabular}%
}
\end{table*}

For all scenarios, $\mathcal{A}=[-1,1]^4$. Under the linear-velocity profile, the first three action components define the desired direction and the fourth scales the speed limit. Under the body-rate and thrust profile, the first component controls collective thrust and the remaining components scale roll, pitch, and yaw rates. The rate limits are $(5,5,4.5)$, $(6,6,5)$, and $(7,7,6.5)$ rad/s for racing, flip traversal, and cluttered navigation, respectively, with corresponding thrust ranges of $[0,2.5]g$, $[0,3]g$, and $[0,2.5]g$.

GPT-5.6-sol with medium reasoning is used as the primary agent for reward generation, behavioral analysis, and refinement, conditioned on the task description, scene image, and observation specification. During deployment, a lightweight VLM agent identifies the observed scenario and selects the corresponding trained policy from the scenario registry.

Real world experiments are conducted on a physical quadrotor using ROS2, an Orange Pi 8GB onboard computer, and an Intel RealSense camera, with Vicon motion capture providing pose measurements for evaluation. The learned policies are validated across two autopilot stacks: ArduPilot, where velocity commands are transmitted through MAVROS at 25~Hz, and PX4, where body-rate and collective-thrust policies are tested in a Gazebo PX4 SITL environment using the uXRCE-DDS bridge and Micro XRCE-DDS agent at 60~Hz. This validates that the same simulation and training framework can support policies deployed through both higher level velocity control and lower level body-rate/thrust control interfaces.

\subsection{Evaluation Metrics}
\label{sec:evaluation-metrics}

We report simulation success rate (\textit{SSR}), real world success rate (\textit{RSR}), sim-to-real accuracy (\textit{S2R}), and reward refinement gain (\textit{RRG}). All eight scenarios are evaluated over 100 simulation trials, while the five linear-velocity scenarios additionally use 10 real world trials each.

For a given scenario $s$, the simulation and real world success rates are defined as:
\begin{equation}
\begin{aligned}
\mathrm{SSR}^{(s)} &=
\frac{N_{\mathrm{sim,success}}^{(s)}}
{N_{\mathrm{sim,total}}^{(s)}} \times 100, \\
\mathrm{RSR}^{(s)} &=
\frac{N_{\mathrm{real,success}}^{(s)}}
{N_{\mathrm{real,total}}^{(s)}} \times 100 .
\end{aligned}
\end{equation}

The collective success rate across all evaluated scenarios is computed as:
\begin{equation}
\mathrm{SR}_{\mathrm{collective}} =
\frac{\sum_{s=1}^{S} N_{\mathrm{success}}^{(s)}}
{\sum_{s=1}^{S} N_{\mathrm{total}}^{(s)}} \times 100 ,
\end{equation}
where $S$ denotes the total number of evaluated scenarios.

The simulation-to-real transfer accuracy for scenario $s$ is defined as:
\begin{equation}
\mathrm{S2R}^{(s)} =
\frac{\mathrm{RSR}^{(s)}}
{\mathrm{SSR}^{(s)}} \times 100 .
\end{equation}


Reward refinement gain (\textit{RRG}) measures the absolute increase in simulation success rate between the policies trained using the initial and final refined reward functions. For scenario $i$, it is defined as:
\begin{equation}
    \mathrm{RRG}_i =
    \mathrm{SSR}^{\mathrm{final}}_i -
    \mathrm{SSR}^{\mathrm{initial}}_i,
\end{equation}
and is reported in percentage points (pp). The collective reward refinement gain is computed as the macro-average across $M$ scenarios:
\begin{equation}
    \mathrm{RRG}_{\mathrm{collective}}
    =
    \frac{1}{M}\sum_{i=1}^{M}\mathrm{RRG}_i.
\end{equation}

\section{Results}
\label{sec:result}

\subsection{Simulation-to-Real and Physical UAV Validation}
\label{sec:sim2real-results}
\begin{figure}[h!]
    \centering
    \includegraphics[width=0.47\textwidth]{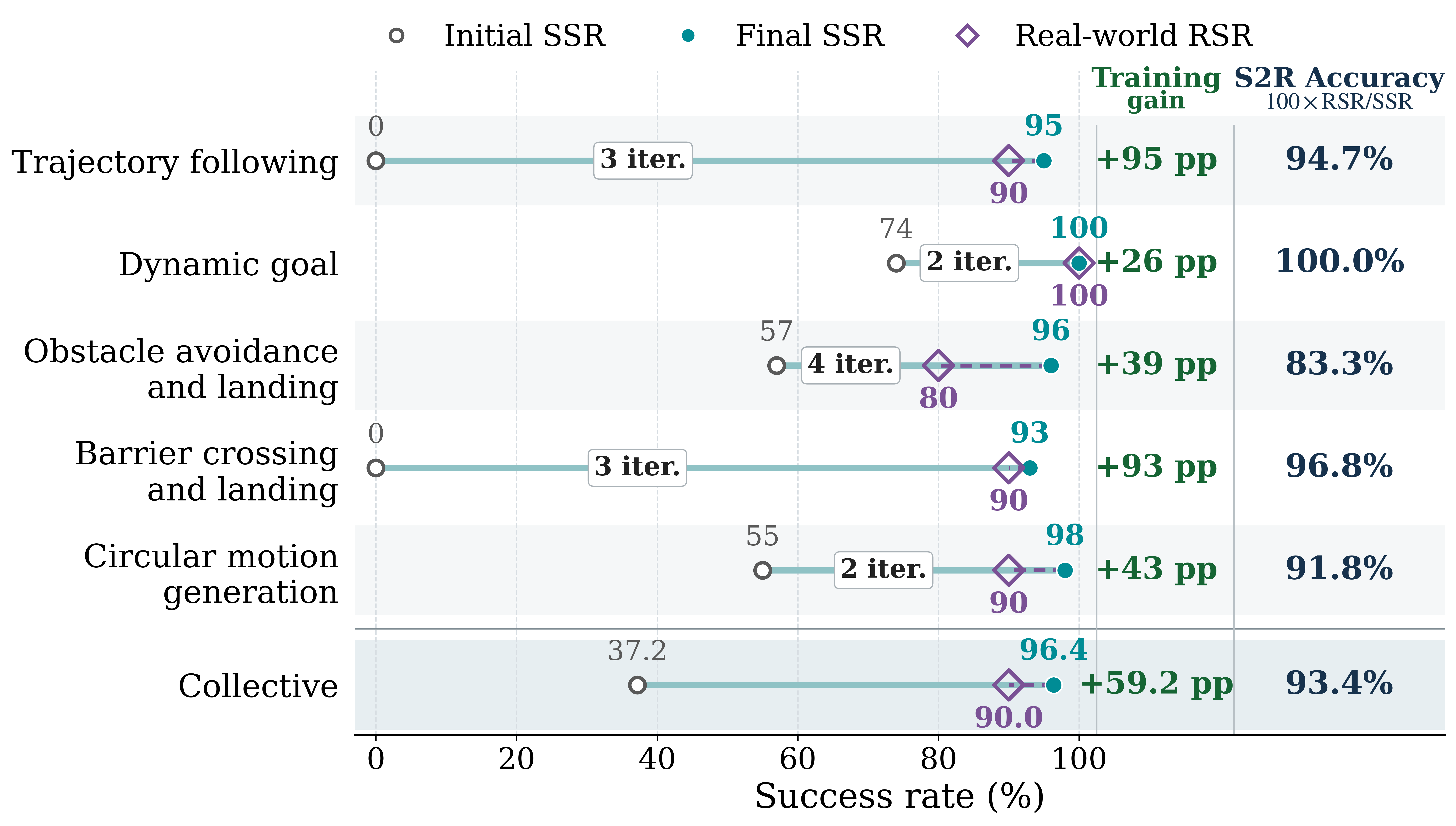}
    \caption{Simulation and real world success rates before and after AgenticRL reward refinement.}
    \label{fig:sim2real}
\end{figure}

Fig.~\ref{fig:sim2real} summarizes reward refinement performance and subsequent transfer to the physical UAV. Across the five real world tested tasks, the mean SSR increases from $37.2\%$ with the initial rewards to $96.4\%$ after refinement, corresponding to a collective RRG of \textbf{$59.2$} percentage points (pp). The refined policies achieve a collective RSR of \textbf{$90.0\%$} and a collective S2R accuracy of \textbf{$93.4\%$}.

The largest refinement gains occur in trajectory following and barrier crossing and landing, where SSR improves by $95$ and $93$ pp, respectively. Dynamic goal, obstacle avoidance and landing, and circular motion generation improve by $26$, $39$, and $43$ pp, respectively. Dynamic goal achieves perfect S2R accuracy, while trajectory following retains $94.7\%$ of its simulation performance. Barrier crossing and landing achieves an RSR of $90\%$ and an S2R accuracy of $96.8\%$. Circular motion generation also achieves an RSR of $90\%$, corresponding to an S2R accuracy of $91.8\%$. Obstacle avoidance and landing exhibits the largest transfer gap, achieving an RSR of $80\%$ and an S2R accuracy of $83.3\%$.

\begin{figure}[h!]
    \centering
    \includegraphics[width=0.40\textwidth]{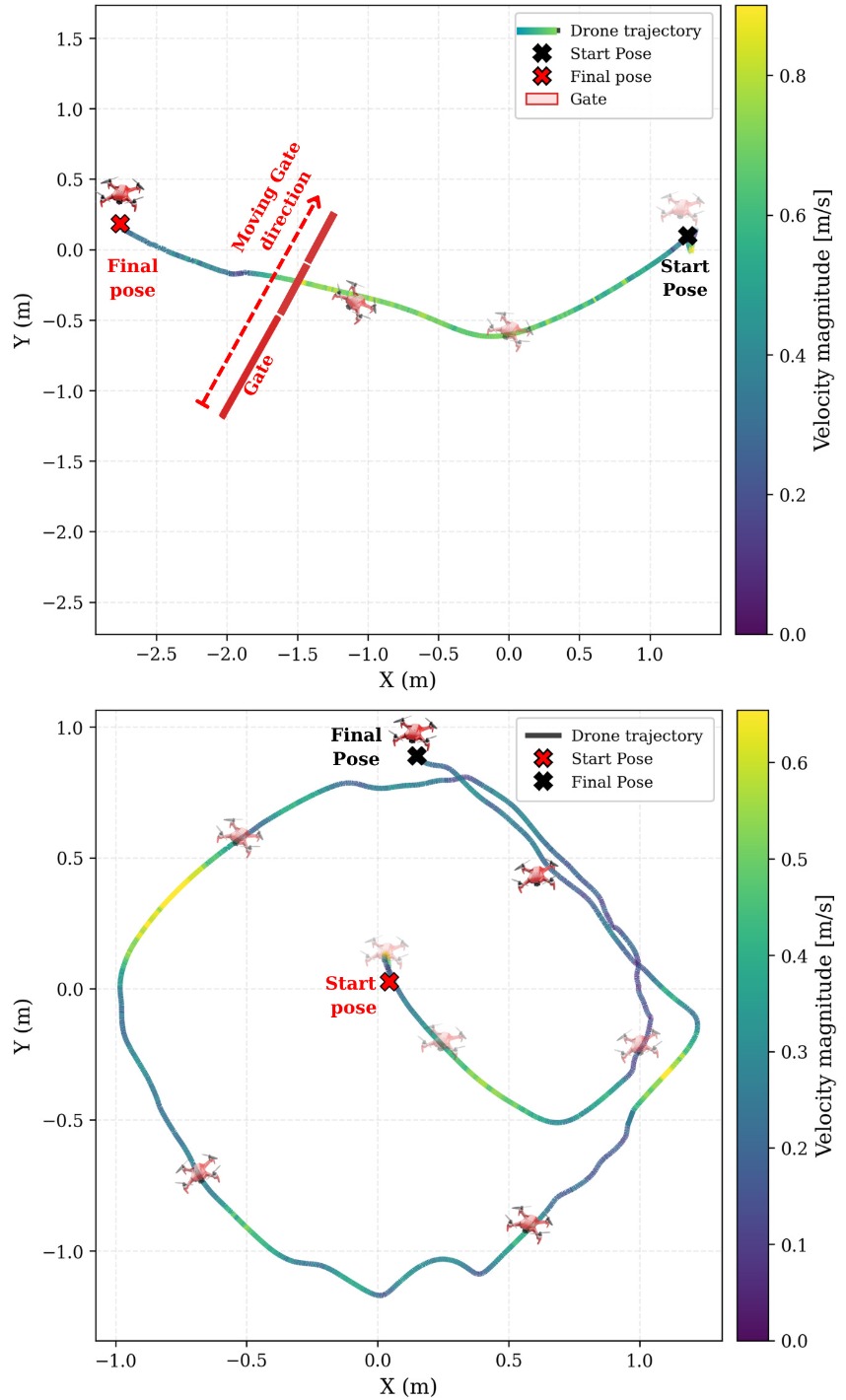}
    \caption{Top-view real world trajectories for dynamic goal and circular motion generation. Color denotes velocity magnitude.}
    \label{fig:realworld_2d_results}
    \vspace{-6pt}
\end{figure}

\begin{figure*}[h!]
    \centering
    \includegraphics[width=\textwidth]{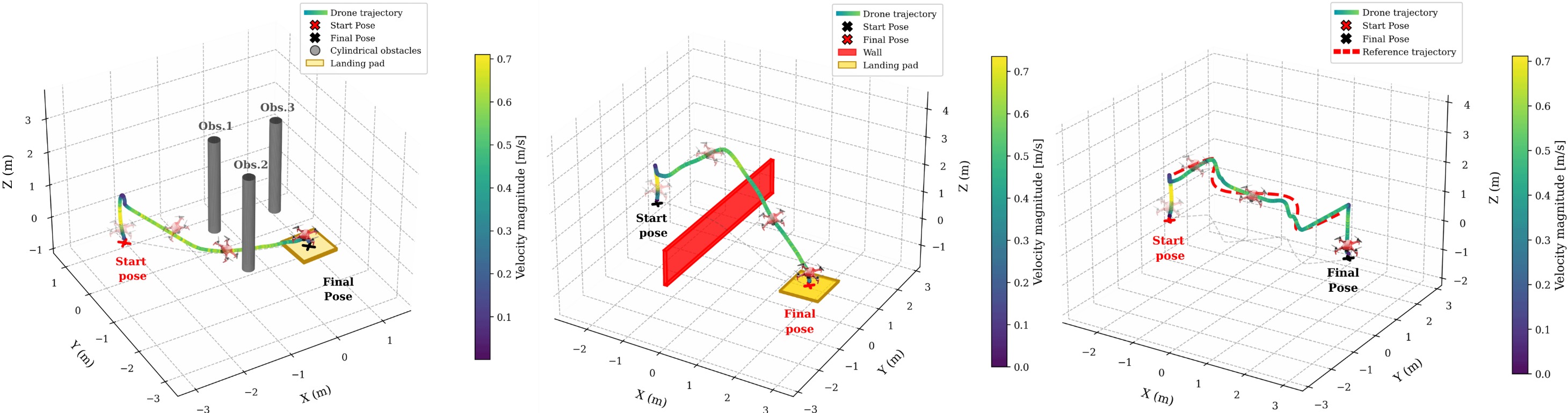}
    \caption{Real world 3D trajectories for obstacle avoidance and landing (left), barrier crossing and landing (middle), and trajectory following (right). Color denotes velocity magnitude.}
    \label{fig:realworld_3d_results}
    \vspace{-0.4cm}
\end{figure*}

Representative physical flight trajectories are shown in Fig.~\ref{fig:realworld_2d_results} and Fig.~\ref{fig:realworld_3d_results}. Fig.~\ref{fig:realworld_2d_results} illustrates dynamic goal and circular motion generation, while Fig.~\ref{fig:realworld_3d_results} shows obstacle avoidance and landing, barrier crossing and landing, and trajectory following. The resulting trajectories demonstrate that the refined policies preserve the intended task behavior after deployment on the physical UAV.

\subsection{Reward Refinement Dynamics}
\label{sec:refinement-results}

AgenticRL refines rewards by using the behavior induced by each generated reward as feedback for subsequent iterations. These revisions are not assumed to be monotonic; therefore, each generated reward is evaluated through the policy behavior it produces, and the best task-valid policy is retained.

To illustrate how the closed loop refinement dynamics operate, Fig.~\ref{fig:reward_revision_example} presents an example from the racing scenario. The initial reward encouraged stable gate traversal but introduced a stationary hovering behavior after the second gate. The diagnosis stage identified this failure by analyzing task-specific behavioral evidence, revealing that the policy satisfied intermediate objectives while failing to complete the overall manoeuvre. Based on this evidence, the refinement agent modified the reward structure by reducing excessive stability incentives and introducing progress, crossing, and anti-stalling terms. The resulting policy improved from $0\%$ to $100\%$ success rate while reducing collision rate and gate-traversal error.

\begin{figure}[h!]
    \centering
    \includegraphics[width=0.5\textwidth]{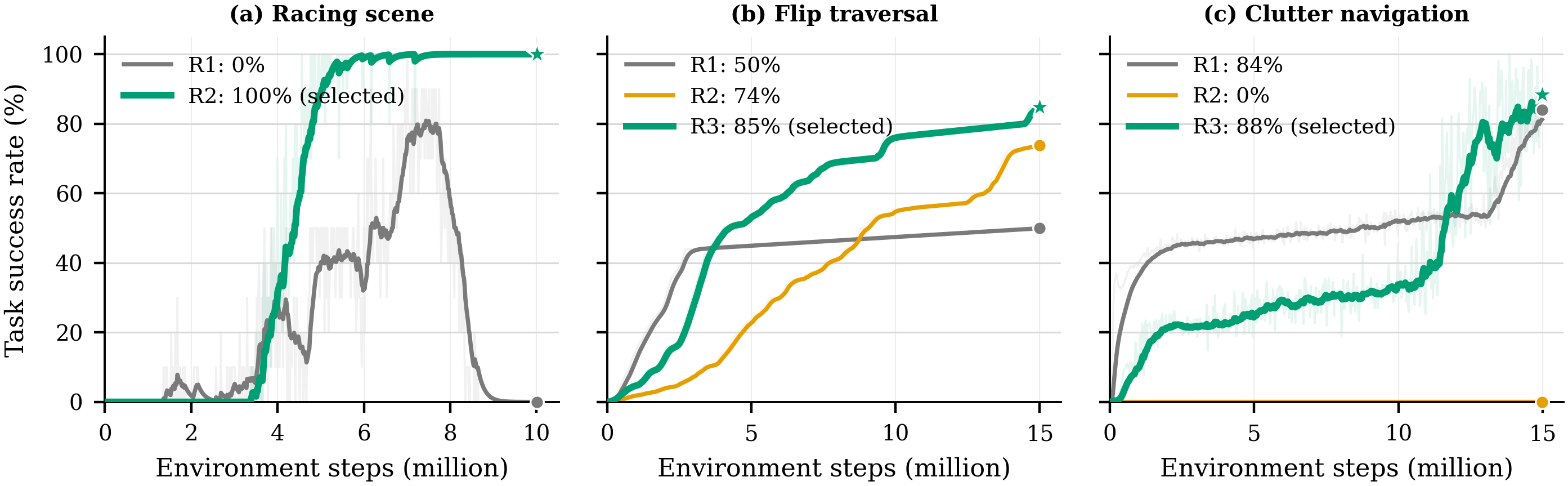}
    \caption{Task success across successive AgenticRL reward iterations for racing, flip traversal, and cluttered navigation. The selected policy is highlighted at its final evaluation value.}
    \label{fig:body_rate_iterations}
\end{figure}

The overall refinement behavior across the three body-rate tasks is shown in Fig.~\ref{fig:body_rate_iterations}. Racing reaches $100\%$ SR after the second reward, flip traversal improves from $50\%$ to $85\%$ over three rewards, and clutter reaches $88\%$ for the selected third reward despite an unsuccessful intermediate refinement. These results indicate that AgenticRL can improve policies through targeted reward revisions guided by task-specific behavioral evidence, rather than relying only on broad candidate generation or parallel reward search. By identifying failure modes and adapting the reward accordingly, the framework focuses refinement on the behaviors required for task completion.

\definecolor{revblue}{HTML}{0072B2}
\definecolor{revpurple}{HTML}{7B61A8}
\definecolor{revgray}{HTML}{555555}
\definecolor{revteal}{HTML}{00897B}
\begin{figure*}[h!]
\centering
\makebox[\textwidth][c]{%
%
\begin{minipage}[c]{0.15\textwidth}
\begin{tcolorbox}[
    enhanced,
    title={\scriptsize Behavioral evidence},
    fonttitle=\bfseries,
    coltitle=white,
    colback=revblue!4,
    colframe=revblue,
    colbacktitle=revblue,
    boxrule=0.45pt,
    arc=1mm,
    left=0.7mm,
    right=0.7mm,
    top=0.5mm,
    bottom=0.5mm,
    toptitle=0.4mm,
    bottomtitle=0.4mm,
    height=26mm,
    valign=center,
    halign=left]
\scriptsize
Both gates: $94/100$\\
Stable exit: $0/100$\\
Exit distance: $1.306$ m\\
Behavior: stationary hover
\end{tcolorbox}
\end{minipage}
\hspace{0.004\textwidth}
%
\begin{minipage}[c]{0.026\textwidth}
\centering
{\normalsize\textcolor{black!60}{$\longrightarrow$}}
\end{minipage}
\hspace{0.004\textwidth}
%
\begin{minipage}[c]{0.165\textwidth}
\begin{tcolorbox}[
    enhanced,
    title={\scriptsize Agent diagnosis},
    fonttitle=\bfseries,
    coltitle=white,
    colback=revpurple!4,
    colframe=revpurple,
    colbacktitle=revpurple,
    boxrule=0.45pt,
    arc=1mm,
    left=0.7mm,
    right=0.7mm,
    top=0.5mm,
    bottom=0.5mm,
    toptitle=0.4mm,
    bottomtitle=0.4mm,
    height=26mm,
    valign=center,
    halign=left]
\scriptsize
Persistent stability rewards create a profitable stationary equilibrium after the second gate.
\end{tcolorbox}
\end{minipage}
\hspace{0.004\textwidth}
%
\begin{minipage}[c]{0.026\textwidth}
\centering
{\normalsize\textcolor{black!60}{$\longrightarrow$}}
\end{minipage}
\hspace{0.004\textwidth}
%
\begin{minipage}[c]{0.30\textwidth}
\begin{tcolorbox}[
    enhanced,
    title={\scriptsize Representative reward code revision},
    fonttitle=\bfseries,
    coltitle=white,
    colback=revgray!3,
    colframe=revgray,
    colbacktitle=revgray,
    boxrule=0.45pt,
    arc=1mm,
    left=0.7mm,
    right=0.7mm,
    top=0.4mm,
    bottom=0.4mm,
    toptitle=0.4mm,
    bottomtitle=0.4mm,
    height=26mm,
    valign=center]
\fontsize{4.8}{5.3}\selectfont\ttfamily
\textcolor{red!75!black}{
- useful\_speed = exp(-((speed - 0.65)/0.32)**2)}\\
\textcolor{red!75!black}{
- reward += stability\_weight * upright\_alignment}\\[0.3mm]
\textcolor{green!40!black}{
+ approach = 1.20 * tanh(approach\_velocity/0.60)}\\
\textcolor{green!40!black}{
+ crossing = 0.85 * near\_plane * tanh(normal\_velocity/0.50)}\\
\textcolor{green!40!black}{
+ stall = -0.22 * tanh(rel\_norm/0.50) * stall\_factor}
\end{tcolorbox}
\end{minipage}
\hspace{0.004\textwidth}
%
\begin{minipage}[c]{0.026\textwidth}
\centering
{\normalsize\textcolor{black!60}{$\longrightarrow$}}
\end{minipage}
\hspace{0.004\textwidth}
%
\begin{minipage}[c]{0.16\textwidth}
\begin{tcolorbox}[
    enhanced,
    title={\scriptsize Retrained policy},
    fonttitle=\bfseries,
    coltitle=white,
    colback=revteal!4,
    colframe=revteal,
    colbacktitle=revteal,
    boxrule=0.45pt,
    arc=1mm,
    left=0.7mm,
    right=0.7mm,
    top=0.5mm,
    bottom=0.5mm,
    toptitle=0.4mm,
    bottomtitle=0.4mm,
    height=26mm,
    valign=center,
    halign=left]
\scriptsize
Success rate: $0\%\!\rightarrow\!100\%$\\
Collision rate: $6\%\!\rightarrow\!0\%$\\
Mean gate error: $0.173\!\rightarrow\!0.125$ m
\end{tcolorbox}
\end{minipage}
}
\vspace{2.5mm}
\caption{Behavior-grounded reward revision in racing. The initial policy hovered before the exit; targeted progress and anti-stalling terms increased SR from $0\%$ to $100\%$ and reduced CR from $6\%$ to $0\%$.}
\label{fig:reward_revision_example}
\end{figure*}

\subsection{Comparative Analysis}
\label{sec:baseline-results}
Table~\ref{tab:baseline_comparison} compares AgenticRL with Text2Reward~\cite{xie2024text2reward} and Eureka~\cite{ma2024eureka} under identical environments, foundation model, PPO settings, and evaluation conditions within each task. Text2Reward generates a single reward with optional human-feedback in the loop, whereas Eureka searches three reward candidates per iteration. AgenticRL instead trains one targeted reward proposal per iteration using task-level behavioral diagnosis.

AgenticRL achieves the highest success rate in both evaluation scenes. In racing, it reaches \textbf{$100\%$} SR with \textbf{$0\%$} collision rate and the lowest gate error of \textbf{$0.116$ m}. In clutter, it achieves \textbf{$88\%$} SR with only $1\%$ collision rate. Eureka obtains slightly higher goal progress and normalized area under the learning curve (NAUC) in clutter, but with a substantially higher collision rate of $38\%$.

The API usage reported in Table~\ref{tab:baseline_comparison} reflects the different refinement strategies of each method. Text2Reward requires fewer input and output tokens because it does not incorporate detailed behavioral diagnosis or visual scene information. In contrast, AgenticRL provides richer multimodal context, including task-level behavioral evidence and policy failure analysis, resulting in higher input token usage. However, this additional context enables targeted reward refinement with fewer PPO trainings compared with candidate-based search approaches. The output token usage remains comparable to Eureka, indicating that the increased input context is primarily associated with evidence gathering and analysis rather than substantially longer reward generation.

\begin{table*}[h!]
\centering
\caption{Comparison of automated reward design methods on racing and cluttered UAV tasks. API usage and search budgets are reported as Racing/Clutter.}
\label{tab:baseline_comparison}
\scriptsize
\setlength{\tabcolsep}{2.5pt}
\renewcommand{\arraystretch}{1.12}
\resizebox{\textwidth}{!}{%
\begin{tabular}{lccccccccccccc}
\toprule
& \multicolumn{4}{c}{\textbf{Racing scene}} &
\multicolumn{4}{c}{\textbf{Cluttered scene}} &
\multicolumn{2}{c}{\textbf{API usage (Racing/Clutter)}} &
\multicolumn{3}{c}{\textbf{Search budget (Racing/Clutter)}} \\
\cmidrule(lr){2-5}\cmidrule(lr){6-9}\cmidrule(lr){10-11}\cmidrule(lr){12-14}
\textbf{Method} &
\textbf{SR $\uparrow$} & \textbf{CR $\downarrow$} & \makecell{\textbf{Gate error (m) $\downarrow$}} &
\textbf{NAUC $\uparrow$} &
\textbf{SR $\uparrow$} & \textbf{CR $\downarrow$} & \makecell{\textbf{Goal progress}\\\textbf{($\%$) $\uparrow$}} &
\textbf{NAUC $\uparrow$} &
\makecell{\textbf{Input}\\\textbf{tokens (k)}} &
\makecell{\textbf{Output}\\\textbf{tokens (k)}} &
\makecell{\textbf{Reward}\\\textbf{iterations}} &
\makecell{\textbf{PPO}\\\textbf{trainings}} &
\makecell{\textbf{Environment}\\\textbf{steps (M)}} \\
\midrule
Text2Reward$^{\dagger}$ & 0.00\% & 24.00\% & 0.286 & 0.001 & 19.00\% & \textbf{0.00\%} & 10.77 & 0.000 & \textbf{0.43/3.39} & \textbf{4.80/5.91} & {1/3} & {1/3} & \textbf{10/45} \\
Eureka$^{\ddagger}$     & 91.00\% & 9.00\% & 0.231 & 0.120 & 75.00\% & 38.00\% & \textbf{89.71} & \textbf{0.165} & 15.88/30.02 & 13.70/24.17 & 2/3 & 6/9 & 60/135 \\
\rowcolor{gray!20}
\textbf{AgenticRL} & \textbf{100.00\%} & \textbf{0.00\%} & \textbf{0.116} & \textbf{0.579} & \textbf{88.00\%} & 1.00\% & 87.85 & 0.152 & 110.95/403.12 & 9.54/25.61 & 2/3 & 2/3 & 20/45 \\
\bottomrule
\end{tabular}%
}
\vspace{2pt}

\parbox{0.98\textwidth}{\scriptsize
$^{\dagger}$\textbf{Text2Reward-style:} for racing, this method is evaluated in its vanilla one-shot setting, where a single model call receives the task description and observation specification, generates one reward program, and trains one PPO policy. For clutter, Text2Reward is evaluated with three refinement iterations using human feedback.\quad
$^{\ddagger}$\textbf{Eureka-style:} each iteration generates three reward candidates, trains them independently, selects the best evaluated policy, and uses reward statistics and task metrics for reflection.}
\end{table*}

\begin{figure}[h!]
    \centering
    \includegraphics[width=0.6\linewidth]{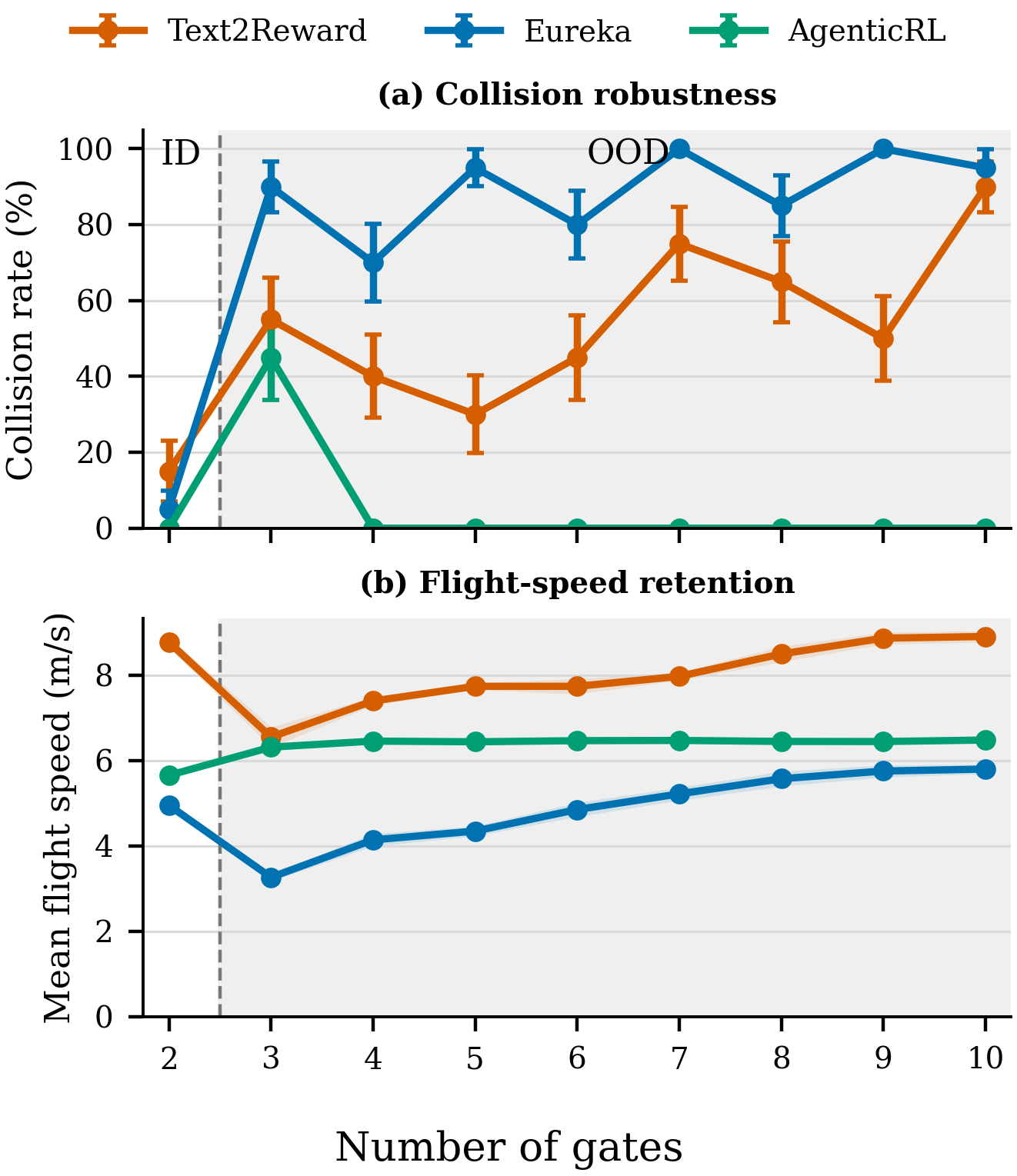}
    \caption{OOD racing generalization from the two-gate training condition to height-varying 3--10 gate courses without retraining. Error bars and bands denote standard error over 20 matched seeds.}
    \label{fig:racing_ood}
\end{figure}

We further evaluate the racing policy, which is trained to traverse a two-gate course in the nominal environment, without retraining by increasing the course complexity. The Out-of-Distribution (OOD) courses contain 3--10 gates arranged in closed loops with a smooth $0.25$ m height variation and independently seeded $\pm0.10$ m per gate height perturbations. Fig.~\ref{fig:racing_ood} reports collision rate and mean flight speed over 20 matched seeds per method and condition.
AgenticRL maintains zero collision rate across all OOD course lengths while preserving a stable flight speed, demonstrating robust generalization beyond the two-gate training condition. In contrast, Text2Reward and Eureka exhibit increased collision rates as the number of gates increases, despite retaining higher flight speeds.

\subsection{Foundation Model Sensitivity}
\label{sec:model-results}
Table~\ref{tab:model_comparison} evaluates the effect of the underlying foundation model within the AgenticRL pipeline on the racing task. GPT-5.6-sol and Gemini 3.1 Pro achieve the highest performance, reaching $100\%$ success, while GPT-5.6-sol obtains the highest NAUC of $0.579$. Gemini 3.1 Pro reaches the success threshold after the initial reward generation, resulting in a single iteration and lower token usage. The variation across models highlights that the choice of foundation model influences reward quality and refinement efficiency within the same AgenticRL framework. Nevertheless, the pipeline remains applicable across different multimodal foundation models, with performance differences primarily reflected in the generated reward revisions.

\begin{table}[h!]
\centering
\caption{Effect of the foundation model within AgenticRL on the racing scene.}
\label{tab:model_comparison}
\scriptsize
\setlength{\tabcolsep}{3.1pt}
\renewcommand{\arraystretch}{1.12}
\resizebox{\columnwidth}{!}{%
\begin{tabular}{lcccccc}
\toprule
\textbf{Model} &
\makecell{\textbf{Best SR}\\\textbf{($\%$) $\uparrow$}} &
\textbf{NAUC $\uparrow$} &
\makecell{\textbf{Iterations}\\\textbf{$\downarrow$}} &
\makecell{\textbf{Input}\\\textbf{tokens (k) $\downarrow$}} &
\makecell{\textbf{Output}\\\textbf{tokens (k) $\downarrow$}} &
\makecell{\textbf{Cost}\\\textbf{(USD) $\downarrow$}} \\
\midrule
GPT-5.4          & 84.00 & 0.202 & 2 & 113.81 & 4.35 & 0.350 \\
Gemini 3.1 Pro   & \textbf{100.00} & 0.320 & \textbf{1} & \textbf{1.80} & \textbf{3.09} & \textbf{0.041} \\
Claude Opus 5    & 46.00 & 0.187 & 2 & 458.29 & 27.98 & 2.991 \\
\rowcolor{gray!20}
GPT-5.6-sol      & \textbf{100.00} & \textbf{0.579} & 2 & 110.95 & 9.54 & 0.176 \\
\bottomrule
\end{tabular}%
}
\end{table}

\subsection{Ablation Study}
\label{sec:ablation-results}

Table~\ref{tab:ablation} evaluates the contribution of reward initialization, behavioral analysis, and vision conditioning across the five linear-velocity scenarios. The evaluated variants are defined as follows: \textbf{zero-shot} reward generates a single reward directly from the task description without refinement iterations; \textbf{few-shot} reward incorporates reward examples and allows iterative refinement for up to two iterations; \textbf{w/o analyzer} retains iterative reward regeneration but removes analyzer-generated behavioral guidance; \textbf{w/o vision} removes scene images while retaining textual and numerical feedback; and \textbf{Full AgenticRL} uses visual context, behavioral diagnosis, and iterative reward refinement. Zero-shot and few-shot reward generation achieve competitive performance on selected tasks but remain inconsistent across scenarios. Removing the analyzer substantially reduces performance, particularly for trajectory following, dynamic goal, and barrier crossing, highlighting the importance of behavior-grounded refinement. Removing vision conditioning also degrades performance, especially in obstacle avoidance, barrier crossing, and motion generation. Overall, the full AgenticRL framework achieves the highest or joint-highest success rate overall.

\begin{table}[h!]
\centering
\caption{Ablation study of AgenticRL across five scenarios.}
\label{tab:ablation}
\scriptsize
\setlength{\tabcolsep}{4pt}
\renewcommand{\arraystretch}{1.15}
\resizebox{\linewidth}{!}{%
\begin{tabular}{lccccc}
\toprule
\textbf{Method} &
\makecell[c]{\textbf{Trajectory}\\
             \textbf{Follow}\\
             \textbf{SR (\%)}} &
\makecell[c]{\textbf{Dynamic}\\
             \textbf{Goal}\\
             \textbf{SR (\%)}} &
\makecell[c]{\textbf{Obstacle Avoidance}\\
             \textbf{\& Land}\\
             \textbf{SR (\%)}} &
\makecell[c]{\textbf{Barrier Crossing}\\
             \textbf{\& Land}\\
             \textbf{SR (\%)}} &
\makecell[c]{\textbf{Circular Motion}\\
             \textbf{Generation}\\
             \textbf{SR (\%)}} \\
\midrule
\rowcolor{gray!14}
Zero-shot reward
& 0 & 74 & 57 & 0 & 55 \\
Few-shot reward
& 48 & 74 & 88 & 72 & 98 \\
\rowcolor{gray!14}
w/o analyzer
& 50 & 61 & 60 & 52 & 62 \\
w/o vision conditioning
& 87 & 95 & 65 & 48 & 76 \\
\midrule
\rowcolor{gray!22}
\textbf{Full AgenticRL}
& \textbf{95}
& \textbf{100}
& \textbf{96}
& \textbf{93}
& \textbf{98} \\
\bottomrule
\end{tabular}%
}
\end{table}

\section{Limitations \& Future Work}
\label{sec:limitations}
The framework introduces additional computational cost due to iterative policy training and evaluation, as each refinement stage requires a new reinforcement learning cycle before subsequent reward optimization. For complex UAV tasks, a complete few-shot refinement process can therefore require considerable training time.

Furthermore, although reward refinement is automated, some environment-specific components still require manual specification, including termination conditions, observation spaces, safety constraints, and task configurations. Thus, a degree of task customization remains necessary when extending the framework to new environments. Finally, the evaluation is conducted in controlled simulation and real world scenarios. While AgenticRL demonstrates effective sim-to-real transfer, further validation in more diverse and highly dynamic environments remains an important direction for future work.

\section{Conclusion}
In this work, we presented \textbf{AgenticRL}, a multimodal closed loop framework for behavior-grounded reward generation and refinement in UAV reinforcement learning. Rather than relying on one-shot reward generation or broad candidate search, AgenticRL evaluates the behavior induced by each generated reward, organizes task-specific behavioral, geometric, and safety measurements into a structured diagnosis packet, and uses this evidence to guide targeted reward revisions. The framework additionally performs task grounding to select compatible action, observation, and reward interfaces, and supports automatic policy selection for deployment.


Evaluation across eight UAV tasks demonstrated the effectiveness of AgenticRL across both linear-velocity and body-rate control settings, supporting policies deployed through both ArduPilot and PX4-based control stacks. Across the five real world validated tasks, reward refinement improved the mean SSR from $37.2\%$ to $96.4\%$, with the refined policies achieving a collective RSR of $90.0\%$ and S2R accuracy of $93.4\%$. AgenticRL outperformed existing automated reward-design baselines in comparative evaluations by using policy-level behavioral evidence to guide targeted reward refinement rather than relying only on one-shot generation or candidate search. The framework remained applicable across different foundation models and evaluation settings, demonstrating its flexibility for multimodal reward design. Ablation studies further verified the contribution of behavioral diagnosis and visual conditioning to reliable reward refinement.

\balance

\bibliographystyle{IEEEtran}
\bibliography{IEEEexample} 

\end{document}